%% file: main.tex
\pdfoutput=1

\documentclass[10pt]{article}

\usepackage{genclaw-techreport}

\usepackage[numbers, sort&compress, square]{natbib}
\usepackage{subcaption}
\usepackage{cleveref}

\newcommand{\figpanelsep}[1][1.4ex]{%
  \par\vspace{#1}%
  \noindent\makebox[\linewidth]{%
    \color{gray!45}%
    \leaders\hbox{\rule[0.5ex]{5pt}{0.6pt}\hspace{4pt}}\hfill
    \hbox{}%
  }%
  \par\vspace{#1}%
}

\title{Editable Visual Design}

\hypersetup{
  pdftitle={Editable Visual Design},
  pdfauthor={Junyan Ye, Wei Liu, Dongzhi Jiang, Zichen Wen, HaoDong Li, Zhutao Lv, Jiaxin Lin, Jinhua Yu, Jun He, Zilong Huang, Rui Chen, Weijia Li},
}

\scaimarker{star}{$*$}
\scaimarker{dag}{$\dagger$}

\scaiauthor{1,2,star}{Junyan Ye}
\scaiauthor{1,star}{Wei Liu}
\scaiauthor{4}{Dongzhi Jiang}
\scaiauthor{5}{Zichen Wen}
\scaiauthor{4}{HaoDong Li}
\scaiauthor{2}{Zhutao Lv}
\scaiauthorbreak                 % 手工换行：前 6 位一行，后 6 位一行
\scaiauthor{1}{Jiaxin Lin}
\scaiauthor{2}{Jinhua Yu}
\scaiauthor{2}{Jun He}
\scaiauthor{2}{Zilong Huang}
\scaiauthor{1,dag}{Rui Chen}
\scaiauthor{3,dag}{Weijia Li}

\affiliation{1}{Tencent Hunyuan}
\affiliation{2}{Sun Yat-sen University}
\affiliation{3}{Tsinghua University}
\affiliation{4}{The Chinese University of Hong Kong}
\affiliation{5}{Shanghai Jiao Tong University}
\metadata[GitHub]{\url{https://github.com/yejy53/Editable-Design}}

\reportnote{%
\normalfont
\textsuperscript{*}\,Equal contribution.\\[-1pt]
\textsuperscript{\textdagger}\,Corresponding authors.%
}

\input{sections/0_abstract}

\begin{document}
\maketitle

% ---------------------------------------------------------------------------
% Teaser 首图 —— 紧接摘要之下，必须落在第一页。
%
% 为何不用 figure 环境：浮动体无法被强制留在首页。首页顶部已被标题与摘要框占据，
% [t] 只会把它浮到第 2 页顶；[h] 在空间不足时也会被降级下沉。故改为非浮动排版，
% 用 \captionof 手工加图注（caption 宏包已由 genclaw-techreport.sty 载入，
% 无需额外引入 float 宏包）。\label 必须写在 \captionof 之后才能绑定正确编号。
%
% 副作用：正文引言被整体推到第 2 页开始。这是刻意取舍 —— teaser 占据首页下半部，
% 符合"摘要 + 首图"的技术报告惯例。
%
% 源文件名为 teasor（作者原始命名，非标准拼写 teaser），沿用未改名以免与作者本地的
% 导出流程脱节；LaTeX 标签用规范拼写 fig:teaser。
% ---------------------------------------------------------------------------
% minipage 是必需的：裸用 center 环境时，LaTeX 允许在图与 \captionof 之间断页，
% 实测出现过"图留在第 1 页、图注被推到第 2 页"的割裂。minipage 不可跨页断开，
% 保证图与图注同进同退。
% 宽度取 0.96\linewidth 而非满栏。这是实测扫描出来的最优值，不是估的：
%
%   1.00 / 0.98  掉出第 1 页
%   0.97 / 0.96 / 0.95  ✓ 图注 2 行
%   0.94  掉出第 1 页        <- 注意这里不单调
%   0.93 / 0.92  ✓ 图注 3 行
%
% 0.94 反常是因为图注行数在临界宽度上跳变：0.95 及以上图注排 2 行，0.94 起变 3 行，
% 而 0.94 的图又比 0.93 高，两者叠加就超出了。所以"越窄越容易放下"在这里不成立，
% 调整宽度后必须重新实测，不能想当然。
%
% 取 0.96 而非临界的 0.97：接近满栏、图注 2 行（比 0.92 时的 3 行更紧凑），
% 且离边界留一档余量。若日后摘要长度、作者人数或单位行数变动，需重新核对 ——
% 判据是图与图注**同时**出现在第 1 页。
%
% 已实测排除的一个方案：注释掉 \contribnote（去掉通讯作者脚注）并不能腾出空间，
% 满栏仍会掉到第 2 页。原因是该脚注位于页脚区（footskip），而版心高度由 geometry
% 固定，与页脚有无内容无关。
% 不再套 center 环境：它上下各加一个 \topsep（合计约 0.7cm 的纯浪费），
% 而 minipage 宽度已是 \linewidth、无需再居中。改为 \noindent + 一个可控的 \vspace，
% 省下的空间正好用来容纳作者增至 10 人后多出的那一行。
% 这里不再另加 \vspace：摘要框与本块之间已有 parskip 提供的段间距（约 0.2cm），
% 首页空间已到边界，多一分都放不下。
\noindent
\begin{minipage}{0.97\linewidth}
  \centering
  % 直接使用作者导出的 teasor_hd.pdf，不再经 PNG 转换：该 PDF 内的文字是矢量的
  % （内嵌 Arial Bold 字体），缩放不失真；图片部分为导出时自带的 JPEG，原样保留。
  \includegraphics[width=\linewidth]{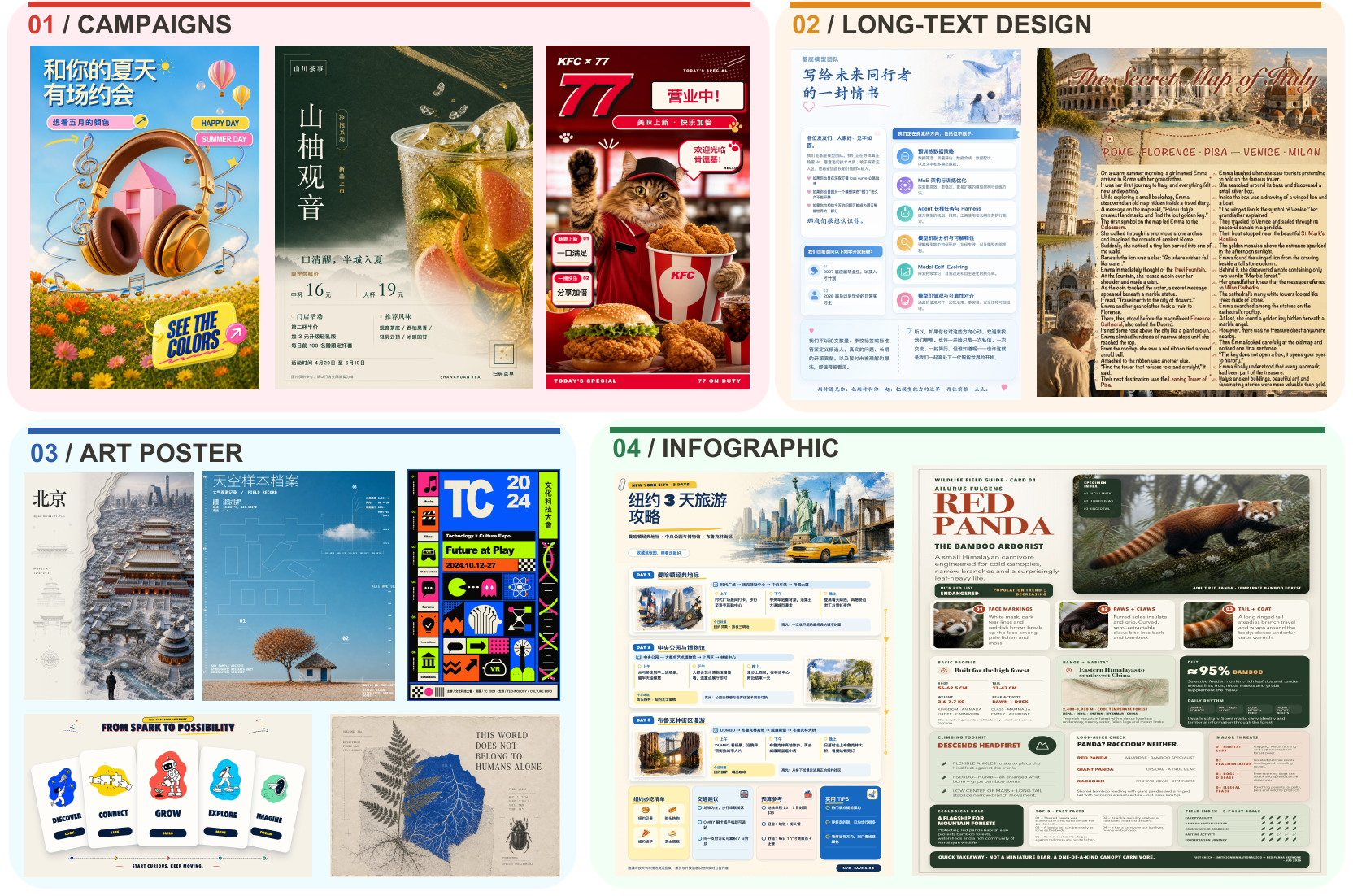}
  % 图注中文版：【图 1】由 Editable Visual Design 生成的可编辑设计产物。
  \captionof{figure}{Editable design artifacts generated by our proposed \textsc{Editable Visual Design}.}
  \label{fig:teaser}
\end{minipage}
\par

\input{sections/1_introduction}

\input{sections/2_workflow}

\input{sections/3_cases}

\input{sections/4_conclusion_discussion}

\input{sections/5_limitations}

\input{sections/6_related_work}

\newpage

\bibliographystyle{plainnat}
\bibliography{ref}

\end{document}

%% file: sections/0_abstract.tex
\begin{abstract}

While diffusion base models such as GPT-Image-2 and Nano-Banana exhibit remarkable visual expressiveness, their end-to-end generation inherently yields flattened bitmaps with error-prone text, precluding layer-wise post-editing. Conversely, code-based visual generation via Coding Agents provides precise layout control and decoupled layers, yet remains constrained by a lack of global aesthetic intuition and the difficulty of coding complex visual assets.

To address this, we propose Editable Visual Design, a new paradigm driven by a Coding Agent. We designate the VLM as the ``creative brain'' for requirement comprehension, task planning, and aesthetic judgment, while utilizing the image generation model as an on-demand ``visual world simulator'' to synthesize standalone visual assets. Operating under an ``imagine first, then act'' closed-loop workflow, the agent generates isolated assets, writes native HTML/CSS, and iteratively refines the design against visual rendering feedback.

Furthermore, Agent Design Replay faithfully reproduces the creative and reasoning trajectory akin to that of professional human designers. Ultimately, the system delivers editable artifacts with decoupled layers and real text, enabling users to perform intuitive mouse dragging and layout adjustments on a graphical user interface. Validations on posters, infographics, and other scenarios show that this paradigm successfully achieves both refined aesthetics and production-grade editability.

\end{abstract}

%% file: sections/1_introduction.tex
% =============================================================================
% 1. Introduction / 引言
% 每段英文正文之前，以注释形式保留对应的中文原文。
% =============================================================================

\section{Introduction}
\label{sec:intro}

% 中文：近年来，大语言模型在代码生成领域取得了突破性进展，尤其是在视觉设计
% （Visual Design）任务上展现出了巨大的潜力。以 GPT-5.6 Sol、Claude Fable 5 和
% Kimi K3 为代表的最新一代模型，已经能够通过直接生成 HTML、SVG 和 CSS，自动化
% 构建结构完整、层次清晰的排版版面。这种基于代码的设计构建方式，具备图层清晰、
% 文字可交互以及天然支持二次修改的工程优势，为自动化海报设计与信息图生成带来了
% 极大的想象空间。
In recent years, large language models have made breakthrough progress in code generation, and they show particularly large potential on Visual Design tasks. The latest generation of models, represented by GPT-5.6 Sol~\citep{openai2026gpt56sol}, Claude Fable 5~\citep{anthropic2026fable5}, and Kimi K3~\citep{kimi2026k3}, can already build structurally complete and clearly organized layouts automatically by directly generating HTML, SVG, and CSS. This way of constructing designs through code brings engineering advantages, including clean layers, interactive text, and natural support for later modification, and it opens up a great deal of room for automated poster design and infographic generation.

% 中文：然而，当视觉代码生成试图向“生产级设计”迈进时，遇到了极难突破的美学直觉与
% 素材瓶颈。首先，现有的代码大模型极度缺乏全局视觉把控力。它们精通语法、DOM 树和
% Flexbox 布局，但却没有二维的空间感和视觉直觉。直接让模型编写排版代码，往往只能
% 产出高度模板化、干瘪的“大标题、卡片、圆角阴影”三件套。模型知道怎么写代码，却不
% 知道怎么写才“好看”，难以跨越从“结构正确”到“视觉高级”的鸿沟。其次，视觉代码在
% 复杂素材的构建上存在天然短板。HTML、SVG 和 CSS 极度擅长精确排版与几何对齐，但
% 直接要求模型用纯代码手绘复杂的视觉素材（如电影感背景、3D 主视觉、自然纹理或复杂
% 插画）成本极高且效果僵硬。这导致过去的代码生成往往只能用简单的几何色块、渐变或
% Emoji 进行占位，最终产物看起来像是一个未完成的半成品。
However, when visual code generation tries to move toward ``production-grade design'', it runs into two bottlenecks that are very hard to break through: aesthetic intuition and assets. First, current code LLMs badly lack global visual control. They are fluent in syntax, the DOM tree, and Flexbox layout, but they have no two-dimensional spatial sense~\citep{si2024design2code, gui2025latcoder, wu2025layoutcoder} or visual intuition~\citep{xiao2026aescoder}. Asking a model to write layout code directly usually yields only the highly templated, thin trio of ``big headline, card, rounded shadow''. The model knows how to write code, but not how to write code that looks good, and it struggles to cross the gap from ``structurally correct'' to ``visually refined''. Second, visual code is inherently weak at building complex assets. HTML, SVG, and CSS are extremely good at precise typography and geometric alignment, but asking a model to hand-draw complex visual assets in pure code, such as a cinematic background, a 3D hero visual, natural textures, or an elaborate illustration, is very costly and the result feels stiff. As a result, past code generation could often only use simple geometric color blocks, gradients, or emoji as placeholders, and the final product looks like an unfinished draft.

% 中文：纵观当前的视觉生成领域，现有的研究主要沿着两条正交的路径展开，但各自都存在
% “偏科”现象。一方面是偏向“左脑”的纯代码生成，现有的 Coding Agent 就像是一个只有
% 左脑的系统，精通逻辑与结构，但由于代码本质上是一维的符号序列，模型极度缺乏二维
% 空间感与美学直觉。另一方面是偏向“右脑”的视觉内容生成（如扩散模型），它压缩了人类
% 数百年的美术先验，能够瞬间生成具备顶级构图、光影与质感的画面；但它存在致命的
% “结构死穴”——缺乏严谨的工程逻辑，生成的像素图（Raster Image）不仅存在文字拼写
% 错误与变形，且元素深度粘连、无法分离为可编辑图层，在文字准确性、局部修改与后期交付
% 上天然不可用，无法作为真正可编辑的设计工程产物。
% 更新（2026-09-02）：此处原作“不可分层”（cannot be separated into layers），与 5.2
% 节所述前沿系统在 layer separation 上的显著进步字面冲突。加限定词改为“无法分离为
% 可编辑图层”后两处一致：分离能力确在进步，但分出来的仍是位图层、其中的文字不可编辑
% —— 这恰是本文的论点，故限定后反而更准确。
Looking across the field of visual generation today, existing research mainly follows two orthogonal paths, and each is lopsided. On one side is ``left-brain'' pure code generation: current Coding Agents are like a system with only a left brain, fluent in logic and structure, but because code is essentially a one-dimensional symbol sequence, these models badly lack two-dimensional spatial sense and aesthetic intuition. On the other side is ``right-brain'' visual content generation such as diffusion models~\citep{rombach2022ldm}, which compress centuries of human artistic priors and can instantly produce images with top-tier composition, lighting, and texture. But they have a fatal structural weakness: they lack rigorous engineering logic. The raster images they generate not only contain misspelled and distorted text~\citep{chen2023textdiffuser, chen2025postercraft}, but also have deeply entangled elements that cannot be separated into editable layers~\citep{jia2023cole, inoue2024opencole, cheng2024graphist}, which makes them inherently unusable for text accuracy, local edits, and downstream delivery, so they cannot serve as a genuinely editable design engineering deliverable.

% 中文：为了打破代码生成的美学瓶颈与像素图像的不可编辑缺陷，本文提出了
% Editable Visual Design，一种由 Coding Agent 驱动的可编辑视觉设计新范式。受世界
% 动作模型（World Action Model, WAM）启发，我们将系统划分为“创作大脑”与“视觉世界
% 模拟器”的协同机制：以 VLM 担任创作大脑，统筹需求理解、设计规划、代码构建与结果
% 判断；将图像生成模型作为随时调用的视觉模拟器，负责将抽象想法快速具象化为直观的
% 视觉效果。
To break through the aesthetic bottleneck of code generation and the non-editability of pixel images, we present \textsc{Editable Visual Design}, a new paradigm for editable visual design driven by a Coding Agent. Inspired by the World Action Model (WAM)~\citep{wang2026wam, yuan2026fastwam}, we split the system into a collaborative mechanism of a \textbf{``creative brain''} and a \textbf{``visual world simulator''}: a VLM acts as the creative brain, coordinating requirement understanding, design planning, code construction, and judgment of results; the image generation model acts as a visual simulator that can be called at any time to quickly turn abstract ideas into concrete visual effects.

% 中文：Agent 遵循“先想象、后行动”的创作闭环：先调用模拟器生成视觉想象图以确立构图、
% 光影与色彩先验，随后自主完成素材生成、原生 HTML/CSS 编写与多轮渲染反思修复。
% 同时，我们引入 Agent Design Replay，完整呈现 Agent 类似于人类艺术家从意图规划、
% 素材生成到代码反思修复的全链路轨迹，使设计过程具备完全的可见性、可追溯性与
% 可复现性。
The agent follows an ``imagine first, then act'' creative loop: it first calls the simulator to generate an imagined visual that establishes priors for composition, lighting, and color, and then independently handles asset generation, native HTML/CSS writing, and multiple rounds of render-and-reflect repair. We also introduce \textbf{Agent Design Replay}, which presents the agent's full trajectory---from intent planning and asset generation to code reflection and repair---much like that of a human artist, so that the design process becomes fully visible, traceable, and reproducible.

% 中文：该系统最终交付的是经过确定性检查与视觉验收的可编辑产物。产物中的文本、素材与
% 排版图层完全解耦，支持用户独立选中、拖拽、编辑与导出。
% 更新（2026-09-02）：原文此处为自造专名 Editable and Quality-Gated Artifacts，
% 现改为普通表述（摘要、引言、结论三处同步）；质量把关的事实描述保留。我们在营销物料、信息图、长文本排版及活动海报等多个典型设计案例上验证了该
% 范式的有效性，展示了兼具高质量美学表现与完全可编辑性的设计产物。
What the system ultimately delivers are editable artifacts that have passed deterministic checks and visual review. Text, assets, and layout layers in the artifact are fully decoupled, so users can select, drag, edit, and export each of them independently. We validate the effectiveness of the paradigm on several typical design cases, including marketing materials, infographics, long-text layout, and event posters, and show design artifacts that combine high-quality aesthetics with full editability.

%% file: sections/2_workflow.tex
% =============================================================================
% 2. Coding-Agent-Driven Design Workflow / 核心工作流
% 每段英文正文之前，以注释形式保留对应的中文原文。
% =============================================================================

\section{Coding-Agent-Driven Design Workflow}
\label{sec:workflow}

\begin{figure}[t]
    \centering
    % 作者导出的矢量版本，沿用其原始文件名以便与导出流程对应。
    % 更新（2026-09-02）：作者重新导出了一版，示例改为「纽约 3 天旅游攻略」海报，第三个
    % 面板由 Code--Asset Co-generation 改名为 Structural Coding & Generation。正文
    % 2.3 的小节标题已同步改名（标题中写作 "and" 而非 "&"，行文惯例，指代同一阶段）。
    \includegraphics[width=\linewidth]{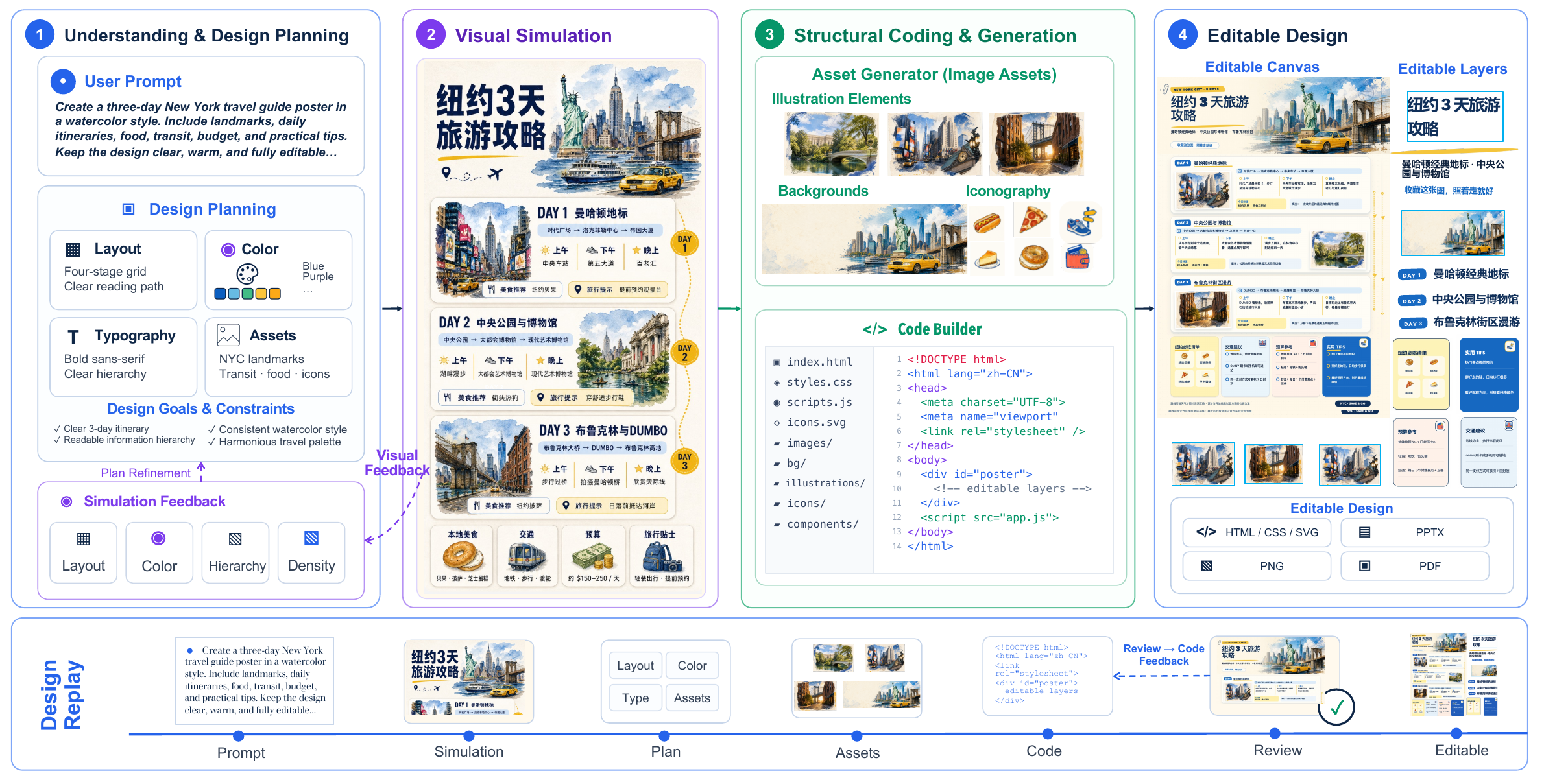}
    % ---------------------------------------------------------------------
    % 图注中文版（中文原稿无对应图注文字，此处为新撰；与下方英文 caption 一一对应）
    %
    % 【图 2】Editable Visual Design 工作流总览。VLM 负责设计规划，调用图像模型取得视觉
    % 想象图与独立素材，再编写原生 HTML/CSS/SVG，最终交付可编辑画布。底部横条是
    % Design Replay，即从需求到产物的完整轨迹记录。
    % ---------------------------------------------------------------------
    % 取舍说明：图上四个面板的名称现已与正文 2.1/2.2/2.3/2.5 的小节标题逐一对应
    % （Understanding & Design Planning、Visual Simulation、Structural Coding &
    % Generation、Editable Design）。唯一不对应的是 2.4 校验 —— 它在图上没有独立面板，
    % 体现为底部 Design Replay 带的 Review 步与 "Review → Code Feedback" 回边。
    % 故图注仍不逐一复述阶段名，改为概述流程走向。
    \caption{\textbf{Overview of the \textsc{Editable Visual Design} workflow.} A VLM plans the design, calls the image model for an imagined visual and for standalone assets, then writes native HTML/CSS/SVG and delivers an editable canvas. The band along the bottom is the \textbf{Design Replay}, the recorded trajectory from prompt to artifact.}
    \label{fig:pipeline}
\end{figure}

% 中文：Editable Visual Design 范式通过多模态大模型（VLM）与图像生成模型的协同，
% 构建从设计需求到可编辑代码产物的完整工作流。系统由担任需求理解、设计规划与排版决策
% 的 VLM（创作大脑）与按需调用的生成模型（视觉模拟器）组成，主要包含理解与设计规划、
% 视觉模拟、结构化编码与生成、校验与视觉自愈以及可编辑设计交付五个步骤。
% 注：句末的图交叉引用为新增（中文原稿无"如图所示"一类表述），属结构性补充而非措辞
% 改动，目的是避免总览图在正文中无处被引用。
% 更新（2026-09-02）：（1）枚举的五步名称已与 2.1--2.5 的小节标题逐字对齐，原先枚举
% 用的是另一套措辞（"视觉模拟、素材生成与排版、渲染校验反思、产物交付"），与小节标题
% 不一致。（2）新增末句披露本文所用的具体模型 —— 此前全文点名了六个他人模型却未说明
% 自身使用什么，使图 3 的对比无法归因。
The \textsc{Editable Visual Design} paradigm builds a complete workflow from design requirement to editable code artifact through the collaboration of a multimodal large model (VLM) and an image generation model~(\Cref{fig:pipeline}). The system consists of a VLM that handles requirement understanding, design planning, and layout decisions (the creative brain) and a generation model that is called on demand (the visual simulator). It has five steps: understanding and design planning, visual simulation, structural coding and generation, verification and visual self-healing, and editable design delivery. In the system reported here, the creative brain is Codex driven by GPT-5.6 Sol~\citep{openai2026gpt56sol}, and the visual simulator is GPT Image 2~\citep{openai2026gptimage2}, which produces both the imagined visual and the standalone assets.

\subsection{Understanding and Design Planning}
\label{sec:workflow_plan}

% 中文（新撰，中文原稿无对应段落）：在生成任何东西之前，Agent 先读需求，明确这件作品
% 要做什么：哪些内容必须出现、交付物是什么形态与尺寸、整体调性该落在哪里。随后它调用
% 图像模型产出一张视觉想象图 —— 那不是结构草图，而是成品可能长什么样的一张画面，作用
% 是为后续流程提供一个足够好的美学参照。这张视觉想象图本身就是设计规划的一部分，而不是
% 规划之前的一步：只有先看到东西，规划才真正定得下来；这里也是用户介入成本最低的位置
% —— 若方向不合心意，可在任何代码产生之前提出并重做。
Before anything is generated, the agent reads the brief and settles what the piece has to do: what content must appear, what the deliverable is and at what size, and what visual register it should sit in. It then calls the image model for an \emph{imagined visual}---not a structural sketch but a picture of what the finished piece could look like, there to give the rest of the process a strong aesthetic reference to work against. This imagined visual is part of the design plan rather than a step before it, since the plan is only settled once there is something to look at; it is also the cheapest place for the user to intervene, saying the direction is not what they wanted and having it redone before any code exists.

\subsection{Visual Simulation}
\label{sec:workflow_sim}

% 中文：在视觉模拟阶段，VLM 对视觉想象图进行视觉解析，提取色彩基调、构图分布以及整体
% 风格特征，作为后续代码构建与视觉设计的全局参考。我们不需要 Agent 对于视觉参考进行
% 一比一复刻，但是通过图像模型获得的结果，还是能够相对比较好的反馈后续 coding agent
% 具体设计执行的阶段，提升其美学与设计感。这个角色和"渲染"是同一类事情：Agent 可以把
% 自己写的代码在浏览器里跑起来、看到刚写出的页面；同样，它也可以调用图像模型，看到一个
% 还没被写出来的设计版本。两者都是把像素交给它去判断，区别在于一个呈现代码当前是什么
% 样，另一个呈现设计可能是什么样。
% 更新（2026-09-02）：（1）"概念草图"统一改称"视觉想象图"（imagined visual）。
% （2）删去原首句"Agent 首先调用图像生成模型产出草图" —— 该动作已移入 2.1，本节只讲
% 如何解析与使用，避免两节重复。（3）末尾两句为新增，用"渲染"类比说明"视觉模拟器"这个
% 称呼的含义，替代在引言里下正式定义的做法。
In the visual simulation stage, the VLM visually parses the imagined visual, extracting the color tone, compositional distribution, and overall style characteristics to serve as a global reference for the code construction and visual design that follow. We do not need the agent to reproduce the visual reference one-to-one; still, the result obtained from the image model gives reasonably good feedback to the coding agent's later stage of concrete design execution, improving its aesthetics and sense of design. The role is the same kind of thing as rendering: the agent can run its own code in a browser and look at the page it just wrote, and it can equally call the image model and look at a version of the design that has not been written yet. Both hand it pixels to judge; one shows what the code currently is, the other what the design could be.

\subsection{Structural Coding and Generation}
\label{sec:workflow_coding}

% 中文：在确立视觉先验后，Coding Agent 承担起核心构图与版面规划的职责。Agent 首先
% 根据视觉先验确定画面的视觉拓扑与图文空间分布，并按需调用生成模型产出干净、独立的
% 局部视觉资产（如无文字背景或独立主体），避免图层粘连与像素污染。随后，Agent 使用
% 原生 HTML/CSS 编写结构化代码，建立清晰的文字排版层级、网格对齐与多图层编排，将
% 解耦的视觉素材与真实文本有机融合为完整且具有层次感的版面。
Once the visual prior is established, the Coding Agent takes on the core responsibility of composition and layout planning. The agent first uses the visual prior to determine the visual topology of the canvas and the spatial distribution of images and text, and calls the generation model on demand to produce clean, standalone local visual assets, such as a text-free background or an isolated subject, which avoids layer entanglement and pixel contamination. The agent then writes structured code in native HTML/CSS, establishing a clear typographic hierarchy, grid alignment, and multi-layer arrangement, and combines the decoupled visual assets with the real text into a complete, well-layered page.

% 中文（新撰，中文原稿无对应段落）：实际执行中，独立素材是逐层单独生成的，而非从视觉
% 想象图上抠取，因此想象图的像素不会进入最终产物。依主体不同走两条路径：模型支持时直接
% 请求带 alpha 通道的素材；否则由提示词把主体置于纯绿背景上，再由抠图脚本取出 —— 只要
% 主体本身不含绿色即可。代码一侧，Agent 直接编写原生 HTML 与 CSS，把页面声明为一块固定
% 像素尺寸的画布，并给每一个使用者可能想单独移动的元素标上独立图层。布局不允许依赖视口
% ，页面在任何地方打开量出来的尺寸都一致，用户拖拽时的坐标才因此有意义。
%
% 事实来源说明：素材两条路径与所用模型由作者提供；代码侧两句核对自开源仓库的
% check-contract.mjs —— 该脚本要求画布声明 data-canvas-width/height 且与实际渲染
% 尺寸相差不超过 1px（C1），要求每个语义模块带唯一 data-layer-id、判据是"使用者是否
% 会想单独移动它"（C3），并通过换视口重载后比对图层坐标来禁止布局依赖 vw/vh/百分比
% （C2）。原先草拟的"每个元素绝对定位""用户打开的就是同一份文档"两句未获脚本佐证
% （交付前另有 bake.mjs 处理），已删去不写。
In practice the standalone assets are generated separately, layer by layer, rather than cut out of the imagined visual, so none of its pixels reach the deliverable. Two routes are used depending on the subject: where the model supports it, the asset is requested directly with an alpha channel; otherwise the prompt places the subject on a flat green background and a matting script lifts it out, which works as long as the subject itself contains no green. On the code side the agent writes native HTML and CSS, declares the page as a canvas of fixed pixel size, and tags every element a user might want to move on its own as a separate layer. Layout is not allowed to depend on the viewport, so the page measures the same wherever it is opened, which is what makes the coordinates a user drags meaningful.

\subsection{Verification and Visual Self-Healing}
\label{sec:workflow_verify}

% 中文：为解决初次生成的代码可能存在的样式溢出、元素重叠或图文遮挡等瑕疵，工作流引入
% 了双重校验与迭代自愈机制。系统首先在无头浏览器环境中加载代码，执行确定性的排版规则
% 检查，检测元素是否存在尺寸溢出、外部资源加载异常或 DOM 结构错误。随后，系统将页面
% 的实际渲染截图输入给 VLM 审查员进行多模态视觉反思。VLM 对比渲染效果与初始设计意图，
% 评估画面的视觉平衡、对齐精度以及文字可读性。若发现瑕疵，Agent 会针对性地生成局部的
% 代码微调补丁，经过一至两轮的迭代反思修复，确保最终渲染效果达到预期质量标准。
To handle flaws that the first pass of generated code may contain, such as style overflow, overlapping elements, or occlusion between images and text, the workflow introduces a dual verification and iterative self-healing mechanism. The system first loads the code in a headless browser environment and runs deterministic layout rule checks, detecting whether elements overflow their size bounds, whether external resources fail to load, or whether the DOM structure is malformed. The system then feeds an actual rendered screenshot of the page to a VLM reviewer for multimodal visual reflection~\citep{yang2024idea2img, yang2025ui2coden, jiang2025draco}. The VLM compares the rendered result against the original design intent and assesses the visual balance, alignment precision, and text readability of the page. When a flaw is found, the agent generates a targeted local patch to fine-tune the code, so that after one or two rounds of reflection and repair the final render reaches the expected quality standard.

\subsection{Editable Design and Agent Design Replay}
\label{sec:workflow_delivery}

% 中文：该工作流最终交付的是具备工程可用性与过程可见性的完整成果。产物本身由原生
% DOM 节点构成，实现了文本、素材与背景图层的完全解耦，用户可以随时进行直接的双击
% 文字编辑、素材拖拽缩放以及分层导出。与此同时，系统将 Agent 从需求拆解、视觉想象、
% 素材提示词、代码演进到最终反思修复的完整决策过程序列化沉淀为 Agent Design Replay。
% 这一设计轨迹不仅使复杂的生成过程告别了传统的不可解释黑盒，也为用户理解设计意图、
% 进行人工介入与二次微调提供了清晰的追溯依据。
What this workflow finally delivers is a complete result that is both usable in engineering terms and transparent in process. The artifact itself is made of native DOM nodes, fully decoupling text, assets, and background layers, so the user can at any time double-click to edit text directly, drag and scale assets, and export layers separately. At the same time, the system serializes the agent's full decision process, from requirement decomposition, visual imagination, and asset prompts to code evolution and final reflective repair, into an \textbf{Agent Design Replay}. This design trajectory not only moves the complex generation process away from the traditional uninterpretable black box~\citep{ye2026genclaw}, but also gives users a clear basis for tracing design intent, intervening manually, and making further adjustments.

\begin{figure}[b]
    \centering
    \includegraphics[width=\linewidth]{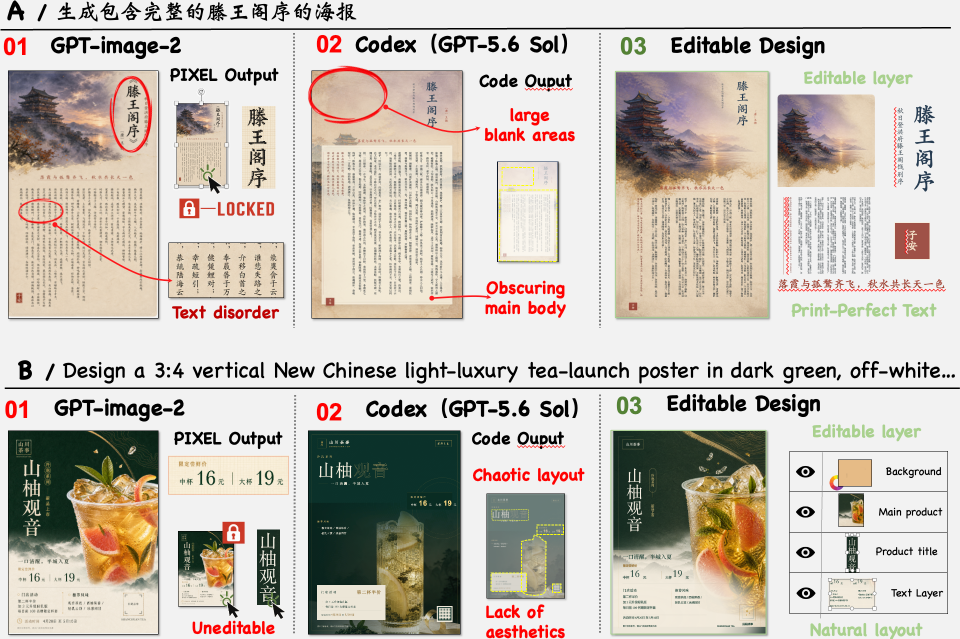}
    % ---------------------------------------------------------------------
    % 图注中文版（中文原稿无对应图注文字，此处为新撰；与下方英文 caption 一一对应）
    %
    % 【图 3】同一需求下与两种范式的对比。GPT-image-2 交付的是锁死的位图，且汉字错乱；
    % 由 GPT-5.6 Sol 驱动的 Codex 交付代码，却留下大面积空白与混乱版式。
    % Editable Visual Design 在保持文字印刷级洁净的同时，交付可分离的图层。
    % 红色标出失败模式，绿色标出分层交付的收益。
    % ---------------------------------------------------------------------
    % 取舍说明：该图把基线具名了（GPT-image-2 与 Codex/GPT-5.6 Sol），故图注顺势引上
    % 书目中已有的对应条目；但 3.1 节正文仍只说"纯扩散模型生图"与"纯 LLM 代码排版"两类
    % 而未点名，正文须忠实中文原稿故未改（见本文件顶部待作者决定事项）。
    \caption{\textbf{Comparison against two paradigms under the same brief.} \emph{GPT-image-2}~\citep{openai2026gptimage2} returns a locked bitmap whose Chinese characters come out disordered; \emph{Codex} on GPT-5.6 Sol~\citep{openai2026gpt56sol} returns code but leaves large blank areas and a chaotic layout. \textsc{Editable Visual Design} keeps the typography print-clean and delivers the page as separable layers. Red marks the failure modes, green what layering adds.}
    \label{fig:comparison}
\end{figure}

%% file: sections/3_cases.tex
% =============================================================================
% 3. Case Studies and Showcase / 案例展示与效果分析
% 每段英文正文之前，以注释形式保留对应的中文原文。
% 插图状态（2026-08-29）：本节三张图全部就位，已无占位符。
%   3.1 基线对比        figure/result1.pdf
%   3.2 多场景展示      figure/result2.pdf
%   3.3 Agent Design Replay，两个案例上下堆叠为一图，中间以 \figpanelsep 浅灰虚线分隔
%       figure/Agent-Replay.pdf（小熊猫）+ figure/Agent-Replay1.pdf（重庆）
%
% 待作者决定（第 2、3 条已于 2026-09-03 关闭，中文原稿与英文正文同步改写）：
%   1. 【未决】3.1 节正文只说"纯扩散模型生图"与"纯 LLM 代码排版"两类而未点名具体系统，
%      但成图已具名 GPT-image-2 与 Codex (GPT-5.6 Sol)。图注已引上对应书目条目。
%      注：CITATION_AUDIT.md §十 记录的"基线未具名"问题 —— 现在图里有具名，2 节也已
%      披露本文所用模型，但 3.1 正文仍未点名，实验协议（数据集/指标/样本量）亦缺。
%      可选修法：在 3.1 加一句"两个基线恰是本系统的两个组件各自单独跑"，把该图变成
%      近似消融的结构。未获作者确认，故未写入。
%   2. 【已关闭】3.2 节原称涵盖"活动海报/信息图表/营销物料/长文本排版"四类而图 4 只有
%      三张海报。核查发现这四类的证据在 teaser（图 1）上，故论断改挂图 1，图 4 改述其
%      真正作用（展开看图层）。论断未删弱，只是指向了能验证它的那张图。
%   3. 【已关闭】3.3 节原只叙述小熊猫一个案例，现已两例并述，并按作者意见把论证重心从
%      图层数量改为"与人类创作流程一致 / 可逐阶段 review / 轨迹可作训练数据"。
% =============================================================================

\section{Case Studies and Showcase}
\label{sec:cases}

% 中文：为了验证 Editable Visual Design 范式的实际表现，我们从基线对比、多场景覆盖
% 以及创作轨迹三个维度展开分析。
To validate how the \textsc{Editable Visual Design} paradigm performs in practice, we analyze it along three dimensions: baseline comparison, coverage of multiple scenarios, and the creative trajectory.

\subsection{Comparative Analysis}
\label{sec:cases_compare}

% 中文：图 1 展示了 Editable Visual Design 与传统“纯扩散模型生图”以及“纯 LLM 代码
% 排版”在相同设计需求下的效果对比。图中可见，纯扩散模型虽然具备良好的画面质感，但
% 生成的文本容易发生形变且图层深度粘连，难以二次编辑；而传统纯代码生成虽结构规整，
% 但因缺乏全局美学先验，产物往往显得扁平单调。相比之下，Editable Visual Design 既
% 借助图像模拟器获得了良好的色彩与氛围质感，又通过代码重构实现了清晰的文字排版与
% 图层解耦，在视觉质感与工程可编辑性之间取得了较好的平衡。
\Cref{fig:comparison} compares \textsc{Editable Visual Design} with conventional ``pure diffusion image generation'' and ``pure LLM code layout'' under the same design requirement. The figure shows that although pure diffusion models produce good visual quality, their generated text is prone to distortion and their layers are deeply entangled~\citep{chen2023textdiffuser, jia2023cole}, making later editing difficult; conventional pure code generation is structurally tidy, but because it lacks a global aesthetic prior, the artifact often looks flat and monotonous. By contrast, \textsc{Editable Visual Design} obtains good color and atmosphere from the image simulator while achieving clear typography and layer decoupling through code reconstruction, striking a good balance between visual quality and engineering editability.

\subsection{Diverse Scenario Showcase}
\label{sec:cases_scenarios}

\begin{figure}[b]
    \centering
    \includegraphics[width=\linewidth]{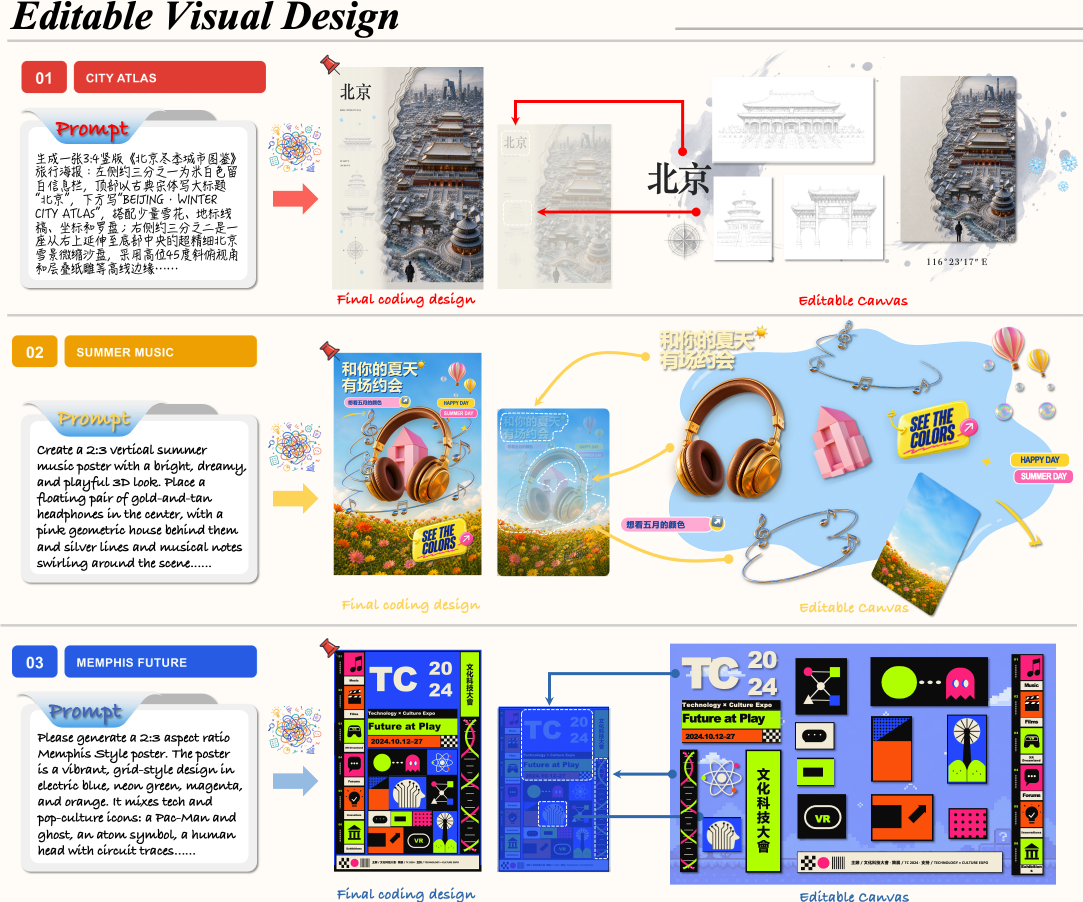}
    % ---------------------------------------------------------------------
    % 图注中文版（中文原稿无对应图注文字，此处为新撰；与下方英文 caption 一一对应）
    %
    % 【图 4】从需求到可编辑画布。三组风格取向各异的需求 —— City Atlas、Summer Music、
    % Memphis Future —— 每行依次为提示词、代码产物，以及同一产物展开后的样子。
    % 右侧面板才是重点：标题字、插画、徽标与背景仍可独立寻址。
    % ---------------------------------------------------------------------
    % 取舍说明：原占位图注沿用正文的"活动海报/信息图表/营销物料/长文本排版"四类，但成图
    % 实际展示的是三种风格迥异的海报，故图注改按图面描述，不再套用那四类，以免与图不符。
    \caption{\textbf{From prompt to editable canvas.} Three briefs in different visual registers---\emph{City Atlas}, \emph{Summer Music}, and \emph{Memphis Future}---each shown as the prompt, the coded result, and the same artifact opened up. The right-hand panels carry the point: headline lettering, illustrations, badges, and background all stay independently addressable.}
    \label{fig:showcase}
\end{figure}

% 中文：图 1 汇总了该工作流在不同设计类型下的产物，涵盖营销物料、长文本排版、艺术海报
% 与信息图表。这些案例显示，该范式能够适应不同信息密度与风格需求：在信息密集场景中保持
% 字阶与网格对齐，在视觉主导场景中合理组织多图层空间。图 4 则把产物展开来看。所有产物
% 均以原生 DOM 形式交付，文字、背景与插画均可在交互视图中独立选中、编辑与导出 —— 看上去
% 是一张压平的海报，实际是一叠可分别寻址的图层。
%
% 修订说明（2026-09-03）：原文把"活动海报、信息图表、营销物料、长文本排版"四类挂在
% 图 4 上，但图 4 实际是三张风格迥异的海报，两者对不上。核查后发现这四类的证据其实在
% teaser（图 1）里 —— 图上明确标着 01 CAMPAIGNS / 02 LONG-TEXT DESIGN /
% 03 ART POSTER / 04 INFOGRAPHIC。故不删这个论断，改为挂到能验证它的那张图上；
% 类别名称也随图面改用图 1 的四个标签。图 4 则改述其真正承担的作用：展开看图层。
\Cref{fig:teaser} collects artifacts this workflow has produced across different design types, covering campaigns, long-text design, art posters, and infographics. These cases show that the paradigm adapts to different information densities and style requirements: in information-dense cases it maintains the type scale and grid alignment, and in visually driven cases it organizes multi-layer space sensibly. \Cref{fig:showcase} then opens the artifacts up. All of them are delivered as native DOM, so text, background, and illustration can each be selected, edited, and exported independently in the interactive view---what looks like a single flat poster is in fact a stack of separately addressable layers.

\subsection{Real-World Case Study of the Agent Design Replay Trajectory}
\label{sec:cases_replay}

% 两个案例合为一图（Figure 4），上下堆叠。
% 实现说明：刻意不预先把两张位图拼成一张 —— 两者内容宽度差约 2%（3659 vs 3583），
% 原尺寸左对齐拼接会在右边缘留下约 3mm 白色锯齿，而右侧恰是深色海报，很显眼。
% 改为各自铺满 \linewidth，缩放交给 PDF 渲染而非提前重采样：画质更好，且两个 PDF
% 都保持逐字节无损。两张源图均已裁去四周白边，使内容跨度可直接比较。
% 面板间距取 1.5ex：足够区分上下两个案例，又不至于割裂成两张图。
% 浮动体放置：必须用 [!t]。两个面板加图注约占 20cm 高，超过 \topfraction 允许的
% 上限，用普通 [t] 会被一路延后到文末（实测被推到第 11 页，而 3.3 节在第 5 页）。
% [!t] 令 LaTeX 忽略高度比例限制，就近放在引用处附近的页顶。
\begin{figure}[!t]
    \centering
    \includegraphics[width=\linewidth]{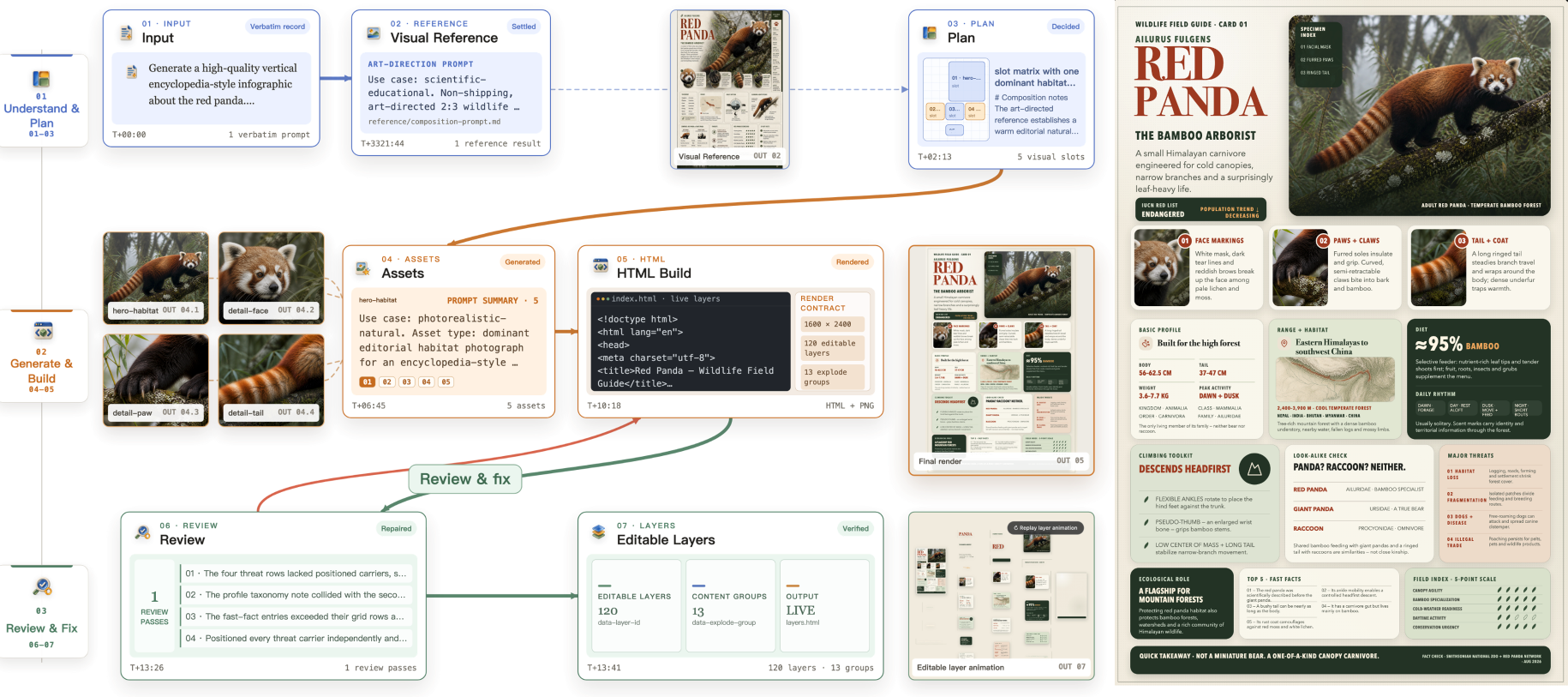}%
    \figpanelsep                                    % 浅灰虚线分隔上下两个案例
    \includegraphics[width=\linewidth]{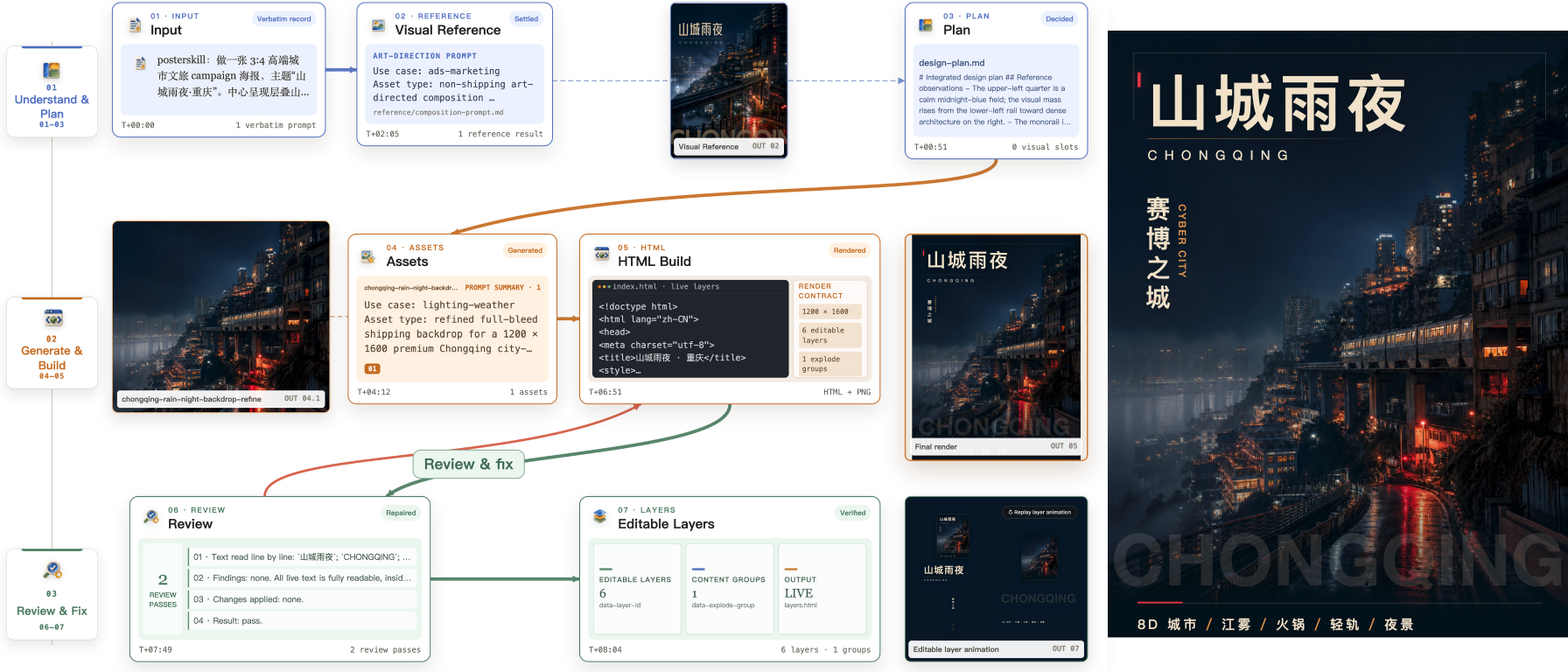}
    % ---------------------------------------------------------------------
    % 图注中文版（中文原稿无对应图注文字，此处为新撰；与下方英文 caption 一一对应）
    %
    % 【图 5】两个真实案例上的 Agent Design Replay。每一步都带时间戳与自己的产出，分三个
    % 阶段：理解与规划、生成与构建、审查与修复。上：信息密集型野外图鉴，13 个分组下
    % 120 个可编辑图层，审查环节查出并修复了真实版式缺陷。下：视觉主导型海报，1 个分组下
    % 6 个图层，审查零改动通过。图层结构随需求的信息密度而变。
    % ---------------------------------------------------------------------
    % 取舍说明一：案例一图中步骤 4 的计数标注为 "5 assets" 但缩略图只画出 4 个，故不写数目。
    % 取舍说明二：案例二 Plan 步的 "visual slots" 计数在图上难以辨认，同样不写具体数字。
    % 取舍说明三：两个案例的对比本身就是论点 —— 信息密集型与视觉主导型 —— 精简后仍保留
    % 图层数与审查结果这两组对照数字，因为它们正是"图层结构随密度而变"的证据。
    \caption{\textbf{Agent Design Replay on two real cases.} Each step carries a timestamp and its own output, across three phases: \emph{understand and plan}, \emph{generate and build}, and \emph{review and fix}. \textbf{Top:} an information-dense field guide---120 editable layers in 13 groups---where the review catches and repairs real layout defects. \textbf{Bottom:} a visually driven travel poster---6 layers in 1 group---where the review passes with no changes. Layer structure follows the density of the brief.}
    \label{fig:replay}
\end{figure}

% 中文：图 5 在两个真实案例上完整呈现了 Agent Design Replay 的创作轨迹。两例中，
% Agent 都是先解析需求、生成视觉想象图，确立版面规划，按需产出独立素材，编写原生
% HTML/CSS 完成多图层构建，随后通过渲染观察反思修复。这样铺开来看，整个序列与人类设计师
% 的工作方式相当一致 —— 先立意图、再具象、备素材、做排版、退后审视、修掉不对的地方 ——
% 这种一致性正是要点所在：过程是可读的，而不是从需求到成图的一次黑盒跳跃。由于每一步都
% 带着自己的产出，轨迹可以逐阶段回看：对结果不满意的人能够判断出是素材生成不够好，还是
% 最后的排版不够好，并直接在那一阶段介入，而不必整体重来。同样的性质也让这些轨迹的价值
% 超出了它们所产出的作品本身 —— 决策内容、决策顺序以及每一步的产出所构成的记录，正是
% 训练一个更强的 Design Agent 所需要的数据。
%
% 修订说明（2026-09-03）：原文只叙述了小熊猫一个案例，而图 5 有小熊猫与旅行海报两个，
% 已改为两例并述。同时按作者意见调整了论证重心：重点不在可编辑图层的数量，而在
% （1）与人类创作流程的一致性、（2）全流程可见、（3）可逐阶段 review 并定点介入、
% （4）轨迹本身可作为训练更强 Design Agent 的数据。图层数与审查结果仍保留在图注里作为
% 图面可验证的对照，正文不再复述。
\Cref{fig:replay} shows the full \textbf{Agent Design Replay} trajectory on two real cases. In both, the agent parses the requirement, generates an imagined visual, settles the layout plan, produces standalone assets on demand, writes native HTML/CSS to build the layers, and then observes the render and reflectively repairs what it finds. Laid out this way the sequence reads much like how a human designer works---form an intent, picture it, gather materials, lay them out, step back and look, fix what is off---and that correspondence is the point: the process is legible rather than one opaque jump from prompt to picture. Because every step carries its own output, the trajectory can be reviewed a stage at a time: someone unhappy with the result can see whether it was the asset generation that fell short or the final layout, and intervene there rather than re-rolling the whole thing. The same property makes these trajectories worth more than the artifacts they produced---a record of what was decided, in what order, and what each decision yielded is the kind of data a stronger design agent could be trained on.

%% file: sections/4_conclusion_discussion.tex
% =============================================================================
% 4. Conclusion and Discussion / 结论与讨论
% 每段英文正文之前，以注释形式保留对应的中文原文。
% =============================================================================

\section{Conclusion and Discussion}
\label{sec:conclusion}

\subsection{Conclusion}
\label{sec:conclusion_main}

% 中文：本文探讨了 Editable Visual Design，一种结合 Coding Agent 与视觉生成模型的
% 可编辑设计探索。通过“VLM 决策规划 + 生成模型视觉模拟”的协同机制，尝试缓解纯代码
% 生成在美学直觉上的不足，并改善传统图像生成中像素不可分层、文字难以修改的问题。在该
% 设计下，Agent 尝试“先借助图像模拟构图与色彩、后执行结构化代码构建”，并通过
% Agent Design Replay 记录从意图规划、素材生成到代码调整的创作过程。最终交付的
% 可编辑产物具备图层解耦与文本可编辑特性，为探索兼顾视觉质感与可维护性的自动化视觉
% 设计提供了一种实践参考。
% 更新（2026-09-02）：自造专名 Editable and Quality-Gated Artifacts 已在摘要、引言、
% 本节三处统一改为普通表述。
This report explores \textsc{Editable Visual Design}, an attempt at editable design that combines a Coding Agent with visual generation models. Through the collaborative mechanism of ``VLM decision planning $+$ generation model visual simulation'', it tries to ease the weakness of pure code generation in aesthetic intuition and to improve on the non-layerable pixels and hard-to-edit text of conventional image generation. Under this design, the agent tries to ``first use image simulation for composition and color, then carry out structured code construction'', and uses \textbf{Agent Design Replay} to record the creative process from intent planning and asset generation to code adjustment. The editable artifacts it finally delivers have decoupled layers and editable text, offering a practical reference for exploring automated visual design that balances visual quality with maintainability.

\subsection{Discussion}
\label{sec:conclusion_discussion}

% 中文：重新审视“生成辅助理解”：从数理求解到视觉灵感构思。长期以来，统一多模态模型
% （UMM）在探索“生成辅助理解（Generation for Understanding）”时，多集中于几何辅助线
% 绘制、空间迷宫等强符号逻辑与线性推理任务，但在这些任务上的尝试往往收益相对有限。
% 究其原因，严密的演绎推理任务可能天然不契合生成模型的隐式反馈机制。一个形象的比喻是：
% 人类很难在梦境中解出严密的数学题，但梦境却常常是视觉灵感、意象与创意构思的来源。
% 扩散模型等生成技术擅长呈现难以用一维纯文本量化的空间美学、色彩氛围与构图参考。本文
% 的探索提示我们，将生成模型前置为视觉模拟器，让 Agent 在编写代码前先借助图像感知全局
% 效果，可能是一种相对自然的生成反哺多模态理解与决策的方式——利用视觉生成获取美学与
% 构图先验，从而在一定程度上辅助后续的代码排版与布局决策。
\noindent\textbf{Revisiting ``Generation for Understanding'': from mathematical and logical problem solving to forming visual ideas.}
For a long time, when unified multimodal models (UMMs)~\citep{deng2025bagel, an2025unictokens} explored ``Generation for Understanding''~\citep{yan2026uae}, they mostly focused on tasks with strong symbolic logic and linear reasoning, such as drawing geometric auxiliary lines or spatial mazes~\citep{qin2026unicot, li2026zebracot}, but the gains from these attempts have often been relatively limited~\citep{shi2026mathcanvas, an2026genius}. The reason may be that rigorous deductive reasoning tasks are inherently a poor fit for the implicit feedback mechanism of generative models. A vivid analogy is this: people can hardly solve a rigorous math problem inside a dream, yet dreams are often a source of visual inspiration, imagery, and creative ideas. Generative techniques such as diffusion models are good at presenting spatial aesthetics, color atmosphere, and compositional references that are hard to quantify in one-dimensional plain text. Our exploration suggests that placing the generation model up front as a visual simulator, letting the agent perceive the overall effect through an image before writing code, may be a relatively natural way for generation to feed back into multimodal understanding and decision-making: using visual generation to obtain aesthetic and compositional priors, and thereby assisting the subsequent code layout and arrangement decisions to some extent.

% 中文：生成模型的分工与协同：作为决策大脑的辅助工具。从 GenClaw、Mind-Brush 到本文
% 的实践，在一定程度上反映出不同模型之间自然的分工定位。将生成模型作为 Agent 随时调用
% 的外挂模拟器与局部素材渲染工具，是一种相对务实且高效的组合方式。在这种分工下，需求
% 拆解、任务规划、代码组织与质量检查主要由具备通用推理能力的 VLM 主导，生成模型则根据
% 需要将抽象构想快速呈现为具体的视觉画面。这种协同方式有助于在发挥生成模型视觉表现力
% 的同时，结合代码实现更精准的结构控制。
\noindent\textbf{Division of labor and collaboration for generation models: a tool that assists the decision brain.}
From GenClaw~\citep{ye2026genclaw} and Mind-Brush~\citep{he2026mindbrush} through to the work in this report, we see, to some degree, a natural division of labor among different models. Treating the generation model as an \textbf{external simulator and local asset renderer that the agent can call at any time} is a relatively pragmatic and efficient combination. Under this division of labor, requirement decomposition, task planning, code organization, and quality checking are mainly led by a VLM with general reasoning ability, while the generation model quickly turns abstract ideas into concrete visuals as needed. This kind of collaboration helps bring out the visual expressiveness of the generation model while using code to achieve more precise structural control.

% 中文：从“位图输出”到“结构化交付”的探索。在实际的设计与应用场景中，设计交付物通常
% 需要具备一定的可维护性与调整空间。传统的文生图模型虽然画面质感丰富，但由于像素深度
% 粘连、文字易错，后期微调较为困难。Editable Visual Design 尝试使用原生 HTML/CSS 和
% 解耦素材来组织版面，使文本、背景与图形图层相对独立，便于用户进行二次选中、修改与
% 导出。这为探索兼顾视觉表现力与确定性可编辑性的设计生成形态提供了一个可行的方向。
\noindent\textbf{Exploring the move from ``bitmap output'' to ``structured delivery''.}
In real design and application settings, a design deliverable usually needs to be reasonably maintainable and to leave room for adjustment. Conventional text-to-image models produce rich visual quality, but because pixels are deeply entangled~\citep{jia2023cole} and text is error-prone~\citep{chen2023textdiffuser}, later fine-tuning is fairly difficult. \textsc{Editable Visual Design} tries to organize the page with native HTML/CSS and decoupled assets, so that text, background, and graphic layers stay relatively independent and users can select, modify, and export them afterwards. This offers a workable direction for exploring forms of design generation that balance visual expressiveness with deterministic editability.

% 中文：Agent Design Replay 与过程可见性：人机协同与未来设计交互的思考。相比于以往
% 直接输出最终结果的黑盒模式，本文通过 Agent Design Replay 记录并呈现 Agent 从需求
% 理解、概念模拟、素材生成到代码调整的过程，有助于提升设计决策链路的透明度与可追溯性。
% 这种过程可见性不仅有助于建立人机协同（Human-in-the-Loop）中的理解与信任，也为未来
% 进一步探索更自然的设计交互形式（例如 AI 结合 GUI 操作鼠标在画布上直接排版与绘制）
% 提供了有益的参考思路。
\noindent\textbf{Agent Design Replay and process visibility: thoughts on human-AI collaboration and future design interaction.}
Compared with the earlier black-box mode that directly outputs a final result, this report uses \textbf{Agent Design Replay} to record and present the agent's process from requirement understanding, concept simulation, and asset generation to code adjustment, which helps improve the transparency and traceability of the design decision chain. This process visibility not only helps build understanding and trust in human-AI collaboration (Human-in-the-Loop), but also offers a useful reference for further exploring more natural forms of design interaction in the future, such as an AI operating a mouse through a GUI to lay out and draw directly on a canvas.

%% file: sections/5_limitations.tex
% =============================================================================
% 5. Limitations / 局限性
%
% 本节为新撰（2026-09-02），中文原稿无对应章节，故中文注释与英文正文同为新写，
% 二者一一对应。四段各自回答一个问题：
%   1. 依赖底座模型  —— 尤其是图像模型的设计感，而非单纯的画质
%   2. 长篇设计      —— 多页/整站的跨页连贯性，对模型能力要求更高
%   3. 审美难以定量  —— 故报告案例而非分数；视觉审查那一半是 VLM 判断
%   4. 可编辑性难以评估 —— 图层数可报，能不能好用不可约化为数字
%
% 已考虑后舍弃的两条（作者判断）：
%   - 绿幕抠图对绿色主体失效：属实但不关键，已在 2.3 节正文以限定从句交代。
%   - “固定像素画布不会重排、换长标题会溢出”：此说不成立（作者确认），故不写。
%     它原出自对 check-contract.mjs 约束的推论，未经实际行为验证。
% 亦按作者要求不设 Ethics 章节。
% =============================================================================

\section{Limitations}
\label{sec:limitations}

% 中文：Editable Visual Design 并不能创造底座模型本身不具备的能力；它所做的是把每一
% 部分工作交给更擅长它的那个模型，因此交付质量同时受制于两者。若 coding agent 写出的
% 版面代码较弱，参照再好，页面也依然弱。更常见的瓶颈在图像模型的设计感：它返回的画面
% 技术上干净，构图却平庸，既没有明确的视觉重心，也没有值得沿用下去的色彩想法 —— 这样
% 一张参照给不了 Agent 多少可用的东西。素材路径另有自己的要求，需要模型能返回与背景
% 干净分离的主体。
\textsc{Editable Visual Design} does not create ability the underlying models lack; what it does is route each part of the job to whichever model is better suited to it, so what gets delivered is bounded by both. If the coding agent writes weaker layout code, the page is weaker however good the reference was. More often the binding constraint is the image model's sense of design: it returns something technically clean but compositionally ordinary, with no clear focal point and no colour idea worth carrying forward, and a reference like that gives the agent little to build on. The asset route makes its own demand, needing the model to return a subject cleanly separated from its background.

% 中文：篇幅更长的设计比单张更难。多页演示或整站页面必须跨页保持一致 —— 同一套字阶、
% 同一组配色、一条能从上一页延续到下一页的视觉线索 —— 而这份一致性要在 Agent 处理一件
% 远长于海报的作品的过程中始终维持住。本文的案例都是单页设计，这一路径在更长篇幅上能
% 走多远，很大程度上取决于底座模型的能力。
Longer pieces are harder than single ones. A multi-page deck or a full website has to stay consistent across pages---the same type scale, the same palette, a visual thread that carries from one page to the next---and that consistency has to be held while the agent works through something far longer than a poster. The cases in this report are all single-page designs, and how far the approach scales at that length depends heavily on the capability of the underlying models.

% 中文：审美质量很难用评估生成质量的常规方式来评估。一张海报“应该”长什么样并没有
% ground truth，而本文所追求的那些性质 —— 构图是否显得经过斟酌、色彩是否有氛围、字阶
% 读起来是否像是有意为之 —— 恰恰是人们会有分歧的东西。因此我们报告案例而非分数；质量
% 检查中视觉的那一半是 VLM 的判断，它替代的是设计师的眼睛，而不是在测量什么。
% 引用说明（2026-09-03）：末句挂 AesEval-Bench —— 该文正是系统评测 VLM 能否评判平面
% 设计审美，结论为存在明显差距，恰好为这句自述提供外部证据，属承重引用而非顺带一提。
Aesthetic quality is hard to evaluate in the way generation quality usually is. There is no ground truth for what a poster should look like, and the properties this work is after---whether a composition feels considered, whether the colour carries atmosphere, whether the type scale reads as deliberate---are exactly the ones people disagree about. We therefore report cases rather than scores, and the visual half of the quality check is a VLM judgment that stands in for a designer's eye rather than measuring anything~\citep{an2026aeseval}.

% 中文：可编辑性同样是断言容易、度量难。图层数与 DOM 结构可以报出来，但真正要紧的是
% 打开这份文件的设计师能不能让它做到自己想要的事，而这一点无法约化为一个数字。
Editability is likewise easier to assert than to measure. Layer counts and DOM structure can be reported, but what matters is whether a designer who opens the file can get it to do what they want, and that does not reduce to a number.

%% file: sections/6_related_work.tex
% =============================================================================
% 6. Related Work / 相关工作
% 编号变更（2026-09-02）：新增 5. Limitations 后本节顺延为第 6 节，文件亦由
% 5_related_work.tex 改名为 6_related_work.tex（git mv，保留历史）。节内 \label
% 未变，交叉引用自动跟随。
% 每段英文正文之前，以注释形式保留对应的中文原文。
%
% 引用说明：中文原稿中的“（作者 et al., 年份）”标注，在此一律替换为对应的 citep
% 命令，key 与 ref.bib 一一对应。ref.bib 中每一条都已比对 arXiv 原文 PDF 首页核验过
% 标题与完整作者名单，核验记录见 CITATION_AUDIT.md。
% 更新（2026-08-27）：GPT-5.6 Sol / Claude Fable 5 / Kimi K3 三个前沿模型原先无引用，
% 现已补上官方来源。三者均经一手来源核验，且 Kimi K3 技术报告正文同时点名另外两个
% 模型，形成交叉印证。详见 CITATION_AUDIT.md。
% =============================================================================

\section{Related Work}
\label{sec:related}

\subsection{Visual Code Generation and LLMs}
\label{sec:related_code}

% 中文：视觉代码生成能力的跃升，最直接地体现在最新一代前沿大模型上。GPT-5.6 Sol、
% Claude Fable 5、Kimi K3 等模型已能根据自然语言指令直接构建交互式界面与结构化排版。
% 围绕这一能力，学术界建立了一系列基准与方法：从早期的 Design2Code（Si et al., 2024）、
% WebSight（Laurençon et al., 2024）与 Web2Code（Yun et al., 2024），到覆盖多交互与
% 复杂框架的 Interaction2Code（Xiao et al., 2025）与 DesignBench（Xiao et al., 2025）。
% 为提升布局还原度，LaTCoder（Gui et al., 2025）提出 Layout-as-Thought 进行分块生成，
% UICopilot（Gui et al., 2025）与 LayoutCoder（Wu et al., 2025）借助 DOM 层级与布局
% 先验引导生成，UI2Code^N（Yang et al., 2025）则将生成建模为“执行—视觉检视—迭代优化”
% 的闭环。在通用网页之外，矢量与结构化设计亦进展迅速：StarVector（Rodriguez et al.,
% 2025）与 OmniSVG（Yang et al., 2025）将 SVG 建模为代码序列生成，InternSVG（Wang et
% al., 2026）打通了 SVG 的理解与生成，AutoPresent（Ge et al., 2025）与 PPTAgent
% （Zheng et al., 2025）以代码动作构建演示文稿，ChartMimic（Yang et al., 2025）与
% ChartCoder（Zhao et al., 2025）专注于图表代码生成。这些工作确立了视觉代码作为可编辑、
% 可验证交付物的优势。然而，现有方法的优化目标普遍集中在语法合规与布局还原上；受限于
% 纯代码对复杂视觉质感的表达瓶颈，高质感背景、自然纹理或复杂插画难以直接用代码手绘，
% 往往退化为简单的色块或渐变占位。Editable Visual Design 建立在这一代码生成能力之上，
% 重点弥补其在全局美学把控与复杂素材生成上的不足。
The leap in visual code generation ability is reflected most directly in the latest generation of frontier large models. Models such as GPT-5.6 Sol~\citep{openai2026gpt56sol}, Claude Fable 5~\citep{anthropic2026fable5}, and Kimi K3~\citep{kimi2026k3} can already build interactive interfaces and structured layouts directly from natural language instructions. Around this ability, the research community has established a series of benchmarks and methods: from the early Design2Code~\citep{si2024design2code}, WebSight~\citep{laurencon2024websight}, and Web2Code~\citep{yun2024web2code}, to Interaction2Code~\citep{xiao2025interaction2code} and DesignBench~\citep{xiao2025designbench}, which cover multiple interactions and complex frameworks. To improve layout fidelity, LaTCoder~\citep{gui2025latcoder} proposes Layout-as-Thought for block-wise generation, UICopilot~\citep{gui2025uicopilot} and LayoutCoder~\citep{wu2025layoutcoder} use DOM hierarchy and layout priors to guide generation, and UI2Code$^N$~\citep{yang2025ui2coden} models generation as a closed loop of ``execute, visually inspect, iteratively refine''. Beyond general web pages, vector and structured design has also advanced quickly: StarVector~\citep{rodriguez2025starvector} and OmniSVG~\citep{yang2025omnisvg} model SVG as code sequence generation, InternSVG~\citep{wang2026internsvg} connects SVG understanding and generation, AutoPresent~\citep{ge2025autopresent} and PPTAgent~\citep{zheng2025pptagent} build presentations through code actions, and ChartMimic~\citep{yang2025chartmimic} and ChartCoder~\citep{zhao2025chartcoder} focus on chart code generation. These works establish the advantage of visual code as an editable, verifiable deliverable. However, the optimization objectives of existing methods generally center on syntactic compliance and layout fidelity~\citep{xiao2026aescoder}; limited by the bottleneck of pure code in expressing complex visual texture, high-quality backgrounds, natural textures, or elaborate illustrations are hard to hand-draw directly in code and often degenerate into simple color blocks or gradients used as placeholders. \textsc{Editable Visual Design} builds on this code generation ability and focuses on making up for its shortcomings in global aesthetic control and complex asset generation.

\subsection{Image Generation and Computational Design}
\label{sec:related_image}

% 中文：与代码路线互补的是基于扩散模型的图像生成路线。Latent Diffusion（Rombach et
% al., 2022）奠定了高质量文生图的基础范式，BAGEL（Deng et al., 2025）、Qwen-Image
% （Wu et al., 2025）与 Emu3.5（Cui et al., 2025）等基础模型进一步提升了生成质量与
% 文字渲染能力（Ye et al., 2025）。其中一条分支专注于照片级真实感：Z-Image（Z-Image
% Team, 2025）以高效架构实现写实生成与中英双语文字渲染，RealGen（Ye et al., 2025）
% 则以合成图像检测器（Ye et al., 2025; Wen et al., 2025; Kang et al., 2025; Guo et
% al., 2026; Lin et al., 2025）作为奖励信号压制生成图像的伪影。面向海报与信息图等
% 视觉设计任务，前沿系统如 GPT Image 1/2（OpenAI, 2026; Yan et al., 2025）、Nano
% Banana 2（Google DeepMind, 2026）与 Seedream 5.0 Pro（ByteDance Seed, 2026）在
% 高密度文字渲染与图层分离方面取得了显著提升，开源的 SenseNova-U1（Diao et al.,
% 2026）亦支持文字密集的信息图生成与编辑。在学术研究方面，TextDiffuser（Chen et al.,
% 2023）通过显式布局改善图内文字渲染，COLE（Jia et al., 2023）与 OpenCOLE（Inoue et
% al., 2024）将平面设计分解为层次化子任务，Graphist（Cheng et al., 2024）输出包含
% 坐标与层序的结构化信息，PosterCraft（Chen et al., 2025）则通过统一框架与强化学习
% 优化海报的美感与文字准确度。这些工作证明了图像模型在构图与美学质感上的强大潜力。
% 然而，这类方法生成的产物本质仍是像素位图（Raster Image），在严谨长文本排版、局部
% 文字修改与后期工程维护上存在固有局限。Editable Visual Design 并不直接将生成的位图
% 作为最终交付物，而是汲取其沉淀的美学构图先验与局部素材渲染能力，作为上游输入反哺
% 后续的结构化代码构建。
Complementary to the code route is the image generation route based on diffusion models. Latent Diffusion~\citep{rombach2022ldm} established the basic paradigm of high-quality text-to-image generation, and foundation models such as BAGEL~\citep{deng2025bagel}, Qwen-Image~\citep{wu2025qwenimage}, and Emu3.5~\citep{cui2025emu35} have further improved generation quality and text rendering ability~\citep{ye2025echo4o}. One branch of this line focuses on photorealism: Z-Image~\citep{team2025zimage} achieves photorealistic generation and bilingual Chinese--English text rendering with an efficient architecture, and RealGen~\citep{ye2025realgen} uses synthetic image detectors~\citep{ye2025loki, wen2025spotfake, kang2025legion, guo2026omniaid, lin2025forensicchat} as a reward signal to suppress artifacts in generated images. For visual design tasks such as posters and infographics, frontier systems such as GPT Image~1/2~\citep{openai2026gptimage2, yan2025gptimgeval}, Nano Banana~2~\citep{google2026nanobanana2}, and Seedream~5.0 Pro~\citep{bytedance2026seedream5} have achieved marked improvements in high-density text rendering and layer separation, and the open-source SenseNova-U1~\citep{diao2026sensenovau1} also supports text-dense infographic generation and editing. On the research side, TextDiffuser~\citep{chen2023textdiffuser} improves in-image text rendering through explicit layout, COLE~\citep{jia2023cole} and OpenCOLE~\citep{inoue2024opencole} decompose graphic design into hierarchical subtasks, Graphist~\citep{cheng2024graphist} outputs structured information containing coordinates and layer order, and PosterCraft~\citep{chen2025postercraft} optimizes poster aesthetics and text accuracy through a unified framework and reinforcement learning. Recent work further suggests that image-editing models can also encode useful visual representations beyond appearance generation~\citep{liu2026opensourceimageeditingmodels}. These works demonstrate the strong potential of image models in composition and aesthetic quality. However, what such methods produce is still essentially a raster image, which has inherent limitations in rigorous long-text layout, local text modification, and downstream engineering maintenance. \textsc{Editable Visual Design} does not take the generated raster image as the final deliverable; instead, it draws on the aesthetic and compositional priors accumulated in this line of work, together with its local asset rendering ability, as upstream input that feeds the subsequent structured code construction.

\subsection{Agentic and Reasoning-driven Design Generation}
\label{sec:related_agentic}

% 中文：视觉生成正从单步黑盒模式逐步走向“先推理规划、后生成执行”的智能体范式。早期
% 工作如 Visual ChatGPT（Wu et al., 2023）与 GenArtist（Wang et al., 2024）探索了
% 多模态工具调度与规划校验，Idea2Img（Yang et al., 2024）通过草图生成与反思迭代优化，
% RPG（Yang et al., 2024）利用多模态模型进行全局区域规划。随后，GoT（Fang et al.,
% 2025）提出生成思维链（Generation Chain-of-Thought），T2I-R1（Jiang et al., 2025）
% 引入双层思维链与强化学习协同，Uni-CoT（Qin et al., 2026）探索了统一多模态推理，
% Mind-Brush（He et al., 2026）将思考、检索与创作深度融合，SCOPE（Ren et al., 2026）、
% GEMS（He et al., 2026）以及 Qwen-Image-Agent（Zhang et al., 2026）进一步丰富了
% 规划、搜索与上下文反馈机制。直接前作 GenClaw（Ye et al., 2026）指出了传统智能体过度
% 依赖最终端到端出图的局限，提出了代码驱动的生成范式，将代码作为可控的中间画布
% （code → image）；与之并行的一条线索把这一思路延伸到了静态图像之外，VideoCoCo
% （Li et al., 2026）以可执行程序作为过程级思维链，使生成过程保持可检视与可控制。
% Editable Visual Design 延续了智能体协同与分工的核心思想，进一步
% 将技术路径演进为以图像生成作为中间素材与前置模拟器、以代码作为最终工程交付物
% （image → code），推动视觉设计从单一的像素位图走向图层解耦、完全可编辑的结构化产物。
Visual generation is gradually moving from a single-step black-box mode toward an agentic paradigm of ``reason and plan first, then generate and execute''. Early work such as Visual ChatGPT~\citep{wu2023visualchatgpt} and GenArtist~\citep{wang2024genartist} explored multimodal tool scheduling and plan verification, Idea2Img~\citep{yang2024idea2img} refines through sketch generation and reflective iteration, and RPG~\citep{yang2024rpg} uses a multimodal model for global region planning. Subsequently, GoT~\citep{fang2025got} proposed the Generation Chain-of-Thought, T2I-R1~\citep{jiang2025t2ir1} introduced a two-level chain of thought coordinated with reinforcement learning, Uni-CoT~\citep{qin2026unicot} explored unified multimodal reasoning, Mind-Brush~\citep{he2026mindbrush} deeply fuses thinking, retrieval, and creation, and SCOPE~\citep{ren2026scope}, GEMS~\citep{he2026gems}, and Qwen-Image-Agent~\citep{zhang2026qwenimageagent} further enrich planning, search, and context feedback mechanisms. The direct predecessor GenClaw~\citep{ye2026genclaw} pointed out the limitation of conventional agents that over-rely on producing the final image end-to-end, and proposed a code-driven generation paradigm that treats code as a controllable intermediate canvas (code $\rightarrow$ image); a parallel line extends this idea beyond still images, where VideoCoCo~\citep{li2026videococo} uses an executable program as a process-level chain of thought that keeps the generation process inspectable and controllable. More recently, AutoDesign~\citep{luo2026autodesign} formulates multimodal design as a long-horizon agentic coding process that iteratively refines editable code-based artifacts. \textsc{Editable Visual Design} continues the core idea of agent collaboration and division of labor, and further evolves the technical route into one that uses image generation as intermediate assets and as an up-front simulator, with code as the final engineering deliverable (image $\rightarrow$ code), pushing visual design from a single raster image toward a layer-decoupled, fully editable structured artifact.